\documentclass[11pt]{article}

\usepackage[preprint]{acl}

\usepackage{times}
\usepackage{latexsym}
\usepackage{hyperref}
\usepackage{url}
\usepackage{algorithm}
\usepackage{algpseudocode}
\usepackage{graphicx}
\usepackage{enumitem}
\usepackage{comment}
\usepackage{booktabs}
\usepackage{multirow}
\usepackage{ragged2e}
\usepackage{subcaption}
\usepackage{amsmath}
\usepackage{amsfonts}
\usepackage{xcolor}
\usepackage{svg}
\usepackage{colortbl}
\definecolor{implicit}{HTML}{8172B3}
\definecolor{explicit}{HTML}{B8963E}
\definecolor{shade}{HTML}{4C72B0}
\usepackage{soul}
\definecolor{wound}{HTML}{4C72B0}
\definecolor{emotion}{HTML}{DD8452}
\definecolor{fear}{HTML}{55A868}
\definecolor{motivation}{HTML}{C44E52}
\newcommand{\qtype}[1]{\colorbox{#1!14}{\textcolor{#1!85!black}{\textsc{\MakeLowercase{#1}}}}}
\newcommand{\phrase}[1]{{\sethlcolor{\phrasecolor!18}\hl{#1}}}
\newcommand{\example}[5]{
  \RaggedRight\qtype{#1}\par\vspace{2pt}
  \textit{#2}\par\vspace{4pt}
  #3\par\vspace{1pt}
  $\rightarrow$~\textbf{\textcolor{#1!85!black}{#4}}
  & \def\phrasecolor{#1}\RaggedRight #5 \\}

  \definecolor{wound}{HTML}{4C72B0}      \definecolor{emotion}{HTML}{DD8452}
\definecolor{fear}{HTML}{55A868}       \definecolor{motivation}{HTML}{C44E52}
\definecolor{neutral}{HTML}{5F5F5F}
\newlength{\barunit}
\newcommand{\nbar}[2][neutral]{\makebox[1.6em][r]{#2}~{\color{#1!70}\rule[-0.1ex]{#2\barunit}{1.4ex}}}
\newcommand{\cat}[2]{{\setlength{\fboxsep}{1pt}\colorbox{#1!18}{\vphantom{Mg}#2}}}

\definecolor{neutral}{HTML}{5F5F5F}
\newcommand{\wc}[1]{\hfill{\color{neutral}\itshape\,#1 words}}

\usepackage[T1]{fontenc}

\usepackage[utf8]{inputenc}

\usepackage{microtype}

\usepackage{inconsolata}

\usepackage{graphicx}

\title{ToMStories: Interpreting Theory-of-Mind in Synthetic Narratives}

\author{Melanie Subbiah \\
  Columbia University \\
  \texttt{m.subbiah@columbia.edu} \\\And
  Chia-En Jacklyn Tsai \\
  Columbia University \\
  \texttt{jacklyntsai1213@gmail.com} \\\And
  Kathleen McKeown\\
  Columbia University\\
  \texttt{kathy@cs.columbia.edu}}

\begin{document}
\maketitle
\begin{abstract}
Understanding people often requires inferring what they feel or want from how they behave rather than from what they state outright. We ask whether language models can make
these inferences, and how they do so. We introduce Theory-of-Mind Stories (ToMStories), an
evaluation dataset of \textasciitilde 6,000 short synthetically generated narrative vignettes, each paired with a multiple-choice or open-ended question
about a character's \textit{emotion}, \textit{fear}, \textit{motivation}, or \textit{emotional wound}.
These psychological states are conveyed \textit{implicitly} (showing), \textit{explicitly} (telling), or with
\textit{conflict} between dialogue and description. Evaluating thirteen open-weight models (1B--72B parameters), we find that the largest
models match humans on multiple choice but not open-ended questions. 
Our analyses point to simplifications models make in both interpretation and generation of implicit psychological states, indicating that synthetic data can help us understand how models perform interpretation, but likely overestimates how well
models interpret humans.
\end{abstract}

\section{Introduction}

Large language models (LLMs) are used today as conversational companions, requiring models to interpret and handle aspects of the user's psychology, such as 
emotions and motives. Reasoning about human psychology requires Theory-of-Mind (ToM) – the ability to infer the beliefs, feelings, and motivations of another person \citep{Premack_Woodruff_1978, wimmer1983beliefs}. Yet prior work shows that feelings, desires, and intentions have received relatively little attention in existing ToM benchmarks \citep{wang2025rethinkingtheorymindbenchmarks}.

\begin{figure}[t]
  \centering
  \includegraphics[width=\columnwidth]{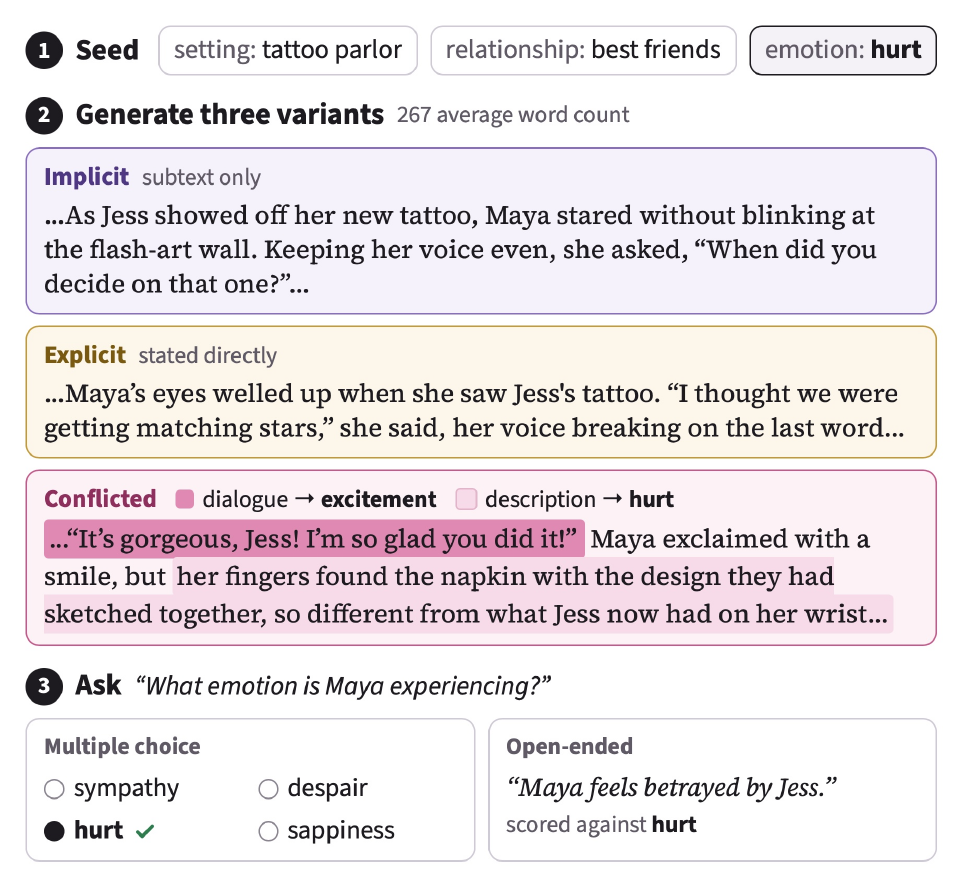}
  \caption{\textbf{ToMStories Pipeline.} (1) For each vignette we sample a setting, relationship, and target psychological state, and (2) generate \textit{implicit}, \textit{explicit}, and
  \textit{conflicted} variants. (3) Models answer multiple choice and open-ended questions about a character's psychology. Open-ended answers are scored against correct answers by embedding-similarity and an LLM judge.} 
  \label{fig:title}
\end{figure}


Narrative can be a fruitful context to study understanding of these aspects of psychology because it often includes subtext and expresses human psychology naturally and realistically. \textit{Subtext} is meaning that is communicated \textit{implicitly}, underlying the \textit{explicit} language. Consider the implicit example shown in Figure \ref{fig:title}. Maya's fixed stare and intentional evening of her voice implies that she is upset. 
Prior work has shown that narrative with psychological subtext like this can be challenging for models to interpret \cite{subbiah-etal-2024-reading}. 

We address this challenge with structured synthetic data that lets us study how models both interpret and generate subtext. Ground-truth interpretations of human-written narrative are hard to establish and often subjective \cite{subbiah2024storysumm, subbiah-etal-2025-top}, so we synthetically generate vignettes that are designed to convey a specific psychological state, giving each a known answer. Each scenario comes in three variants: \textit{implicit}, \textit{explicit}, and \textit{conflicted} (in which dialogue and description suggest different states as shown in Figure \ref{fig:title}). We evaluate thirteen open-weight models on these vignettes, then seek to explain their interpretive performance through analyzing what indicator phrases they associate with different psychological states and how their internal representations capture implicit and explicit expression. 
We focus on open-weight models because they allow for this kind of analysis of internal representations. 


The contributions of our work are\footnote{All data and code is available at: \url{https://github.com/melaniesubbiah/tomstories}.}


\begin{enumerate}[itemsep=1pt, topsep=1pt]
    \item \textbf{ToMStories}, a dataset of synthetically generated narrative vignettes targeting emotions, fears, motivations, and emotional wounds with three variants, two evaluation modes, and human validation. Benchmarking thirteen open-weight models (1B--72B) shows the largest models match humans on multiple choice but fall short on open-ended questions.


    \item \textbf{Evidence that models learn a shared set of indicator phrases to communicate and infer human psychology}. We analyze the overlap in indicator phrases models associate with different psychological states and find that shared indicator phrases with the model used for vignette generation is the most significant predictor of model performance on \textit{implicit} ToMStories.

    \item \textbf{Revealing further simplifications models make in generating and interpreting subtext.} Through analysis of model representations, we find that models treat implicitness partially as muted emotion. Through analysis of performance on \textit{conflicted} ToMStories, we observe that models bias toward dialogue in interpreting meaning, but models that balance dialogue and description more evenly are more accurate on \textit{implicit} ToMStories.

\end{enumerate}

\section{Related Work}
\paragraph{Theory of Mind in LLMs.}
ToM benchmarks for LLMs range from false-belief adaptations \citep{nematzadeh-etal-2018-evaluating,
le-etal-2019-revisiting} to higher-order and multi-party belief tracking \citep{wu-etal-2023-hi,
kim-etal-2023-fantom}, causal templates over beliefs, desires, and percepts
\citep{gandhi2023understandingsocialreasoninglanguage}, narrative ToM linking personality to action
\citep{xu-etal-2024-opentom}, and broad suites of ToM abilities \citep{chen-etal-2024-tombench,
tomato2025}. Yet feelings, desires, and intentions remain underrepresented relative to beliefs
\citep{wang2025rethinkingtheorymindbenchmarks, ma-etal-2023-towards-holistic}. ToMStories targets
this gap with narratives from which a wide range of psychological states must be inferred from
subtext.
\vspace{-0.2cm}
\paragraph{Narrative understanding.}
Models struggle to infer character psychology from implicit language
\citep{subbiah-etal-2024-reading, subbiah2024storysumm}, to track character relationships
\citep{iyyer-etal-2016-feuding}, and, even at long context, to comprehend literature deeply
\citep{thai-iyyer-2025-literary}. Commonsense resources support inference about mental states and
motivations \citep{rashkin-etal-2018-event2mind, sap2019atomicatlasmachinecommonsense}, but over
short event descriptions rather than narratives.
\vspace{-0.2cm}
\paragraph{Mechanistic interpretability.}
Probing \citep{belinkov-2022-probing, hewitt-liang-2019-designing} and steering
\citep{turner2024steeringlanguagemodelsactivation, rimsky-etal-2024-steering, zou2023representation}
are standard tools for reading and manipulating model representations, and probes have decoded
representations of characters' beliefs \citep{3692070.3694663} and emotions
\citep{tak-etal-2025-mechanistic}. Most relevant, \citet{sofroniew2026emotion} find emotion
representations in Claude Sonnet 4.5 that causally shape its outputs. 


\section{ToMStories Dataset}

\begin{table}[t]
\centering \scriptsize
\begin{tabular}{l l >{\raggedright\arraybackslash}p{.5\columnwidth}}
\toprule
\textbf{Category} & \multicolumn{1}{l}{$n$} & \textbf{Examples} \\
\midrule
\multicolumn{3}{l}{\scshape\footnotesize Character} \\
\quad \cat{wound}{Wounds}           & \nbar[wound]{121}      & \textit{infidelity, being bullied, growing up in the shadow of a successful sibling} \\
\quad \cat{emotion}{Emotions}       & \nbar[emotion]{135}    & \textit{anger, giddiness, remorse} \\
\quad \cat{fear}{Fears}             & \nbar[fear]{68}        & \textit{being a burden to others, having no purpose, isolation} \\
\quad \cat{motivation}{Motivations} & \nbar[motivation]{95}  & \textit{being accepted by others, proving someone wrong, protecting a loved one} \\
\midrule
\multicolumn{3}{l}{\scshape\footnotesize Situation} \\
\quad Settings      & \nbar{254} & \textit{abandoned mine, hair salon, hotel room} \\
\quad Relationships & \nbar{48}  & \textit{enemies, nanny and child, people who are dating} \\
\bottomrule
\end{tabular}
\caption{Thesaurus categories used in ToMStories.}
\label{tab:thesaurus}
\end{table}

Our goal is to generate short narrative vignettes that convey a character's psychological state implicitly, while controlling both the scenario and the ground-truth state (pipeline example shown in Figure \ref{fig:title}). We draw scenarios and answers from the Thesaurus Description Database for writers\footnote{\url{https://onestopforwriters.com/thesaurus}. The thesauruses \citep{puglisi2019emotion} guide writers in character development. We use only lists from the free tier and do not reproduce them in our repository; the resource should not be scraped or used commercially.}, using its \textit{emotions}, \textit{fears}, \textit{motivations}, and \textit{emotional wounds} lists as ground-truth answers and its \textit{settings} and \textit{relationships} lists to build scenarios (see summary statistics and examples in Table~\ref{tab:thesaurus}). The \textit{emotions} span a much wider range ($n=135$) than common emotion taxonomies (e.g., $n=28$ in \citet{demszky-etal-2020-goemotions}). For example, the list includes \textit{giddiness} alongside \textit{excitement} and \textit{happiness}. \textit{Fears} are anxieties rooted in a character's experiences, \textit{motivations} are goals a character is driven to pursue, and \textit{wounds} are ways a character has been deeply hurt in the past.

To seed a vignette, we randomly sample a setting, a relationship between two characters, and a psychological state that one of the characters should be experiencing in the scene. Claude Sonnet 4.6 then writes three variants (full prompts in our repository):
\begin{itemize}[itemsep=1pt, topsep=1pt]
\item \textbf{Implicit.} The primary variant conveys the state only through subtext. The model is asked to \textit{write a scenario that does not explicitly mention the <psychological-state> but communicates that one of the characters is experiencing it}, within the given setting and relationship.
\item \textbf{Explicit.} The state is expressed directly: the character \textit{openly and unambiguously shows they are experiencing <psychological-state> through explicit statements, visible physical reactions, and direct behavior}, without using the answer's wording.
\item \textbf{Conflicted.} To test whether models rely more on dialogue or description, a character's \textit{spoken words point to one <psychological-state> while their described behavior and physical reactions reveal a different one}.
\end{itemize}
In each case, the model generates both the vignette and the appropriate corresponding question about the hidden state. For multiple choice questions, we pair the correct answer with three random distractors of the same type in \textit{implicit} and \textit{explicit} vignettes, or the dialogue-aligned and description-aligned answers with two random distractors in \textit{conflicted} vignettes.

\begin{table}[t]
\centering
\scriptsize
\setlength{\tabcolsep}{3pt}
\begin{tabular}{lccc|cccc|c}
\toprule
\textbf{Subset} & \textbf{Tot.} & \textbf{Train} & \textbf{Val.} &
\textbf{Wound} & \textbf{Emot.} & \textbf{Fear} & \textbf{Motiv.} &
\textbf{WC} \\
\midrule
\textit{Implicit}   & 1988 & 1593 & 395 & 493 & 496 & 499 & 500 & 217.4 \\
\textit{Explicit}   & 1988 & 1593 & 395 & 493 & 496 & 499 & 500 & 240.6 \\
\textit{Conflicted} & 1956 & 1561 & 395 & 493 & 493 & 475 & 495 & 342.3 \\
\bottomrule
\end{tabular}
\caption{Summary statistics for ToMStories, showing the number of generated vignettes by split and question type, and the average word count.}
\label{tab:dataset-stats}
\end{table}

\begin{table*}[t]
\centering \scriptsize
\renewcommand{\arraystretch}{1.15}
\begin{tabular}{@{}p{0.31\textwidth} p{0.65\textwidth}@{}}
\toprule
\textbf{Scenario \& question} & \textbf{Vignette excerpt (implicit)} \\
\midrule
\example{wound}{Professional and client at a coffeehouse}
  {What emotional wound does the woman's behavior suggest she has experienced in her past?}
  {being humiliated by others}
  {\ldots She walked him through her finances with quiet precision, answering his questions before he finished asking them, filling every silence before it could stretch. At one point he laughed---warmly, at something she said---and she \phrase{stopped talking mid-sentence, searching his face carefully before deciding it was safe to smile back}. ``You clearly know your numbers,'' he said. ``I just like to be prepared,'' she said, \phrase{straightening papers that didn't need straightening}. ``\phrase{I don't like giving people reasons to}---'' She paused. ``I just like to be prepared.''}
\midrule
\example{emotion}{Police officer and confidential informant at an archery range}
  {What emotion is Marcus feeling?}
  {rage}
  {Marcus pulled the arrow back until the fletching brushed his cheek. \phrase{His jaw was set so tight the muscles in his temple pulsed} visibly beneath the skin. ``You told me the warehouse,'' he said. \phrase{The words came out flat, measured}, like he was reading from a report. He released. \phrase{The arrow punched through the target dead center}, splitting the shaft of the one already there. Detective Yuen took a small step back despite herself\ldots}
\midrule
\example{fear}{Exes at a pool hall}
  {What core fear does Derek demonstrate throughout this scenario?}
  {competition}
  {\ldots Two solids dropped, and she watched him circle the table the way he used to circle every room they walked into together, \phrase{measuring angles, measuring people}. ``You've been practicing,'' she said. ``Little bit.'' He lined up his next shot. She remembered \phrase{the night she'd beaten him at cards in front of his friends. How quiet the drive home had been.} How the silence had its own specific temperature. He sank another ball. Then another. \phrase{He didn't celebrate}, just moved to the next position\ldots}
\midrule
\example{motivation}{Principal and teacher at an outdoor skating rink}
  {What is Principal Dawes's core motivation?}
  {serving others}
  {\ldots Principal Dawes had arrived two hours before the first family showed up, dragging extension cords across the frozen parking lot and \phrase{arranging the hot cocoa station so the shortest kids could reach the ladle}. Now the rink was full, and he moved along its edges like a quiet current. \phrase{He knelt to tighten a kindergartner's skate laces without being asked.} \phrase{He guided an elderly grandfather by the elbow} across a rough patch of ice. When the music cut out, he jogged to the speaker system himself\ldots}
\bottomrule
\end{tabular}
\caption{Example \textit{implicit} vignettes for each question type, highlighting phrases carrying implicit meaning.}
\label{tab:ToMStories_examples}
\end{table*}

We generate 2{,}000 vignettes per variant, ensuring that \textit{implicit} and \textit{explicit} vignettes are paired, sharing the same setting, relationship, and state, so they can be compared directly. After filtering for refusals, answer leakage and keeping pairs intact, we report summary statistics in Table \ref{tab:dataset-stats}. Training and validation splits contain an even balance of the four types of questions. Table~\ref{tab:ToMStories_examples} shows examples.

To verify the vignettes are well-formed, we confirm humans can interpret the \textit{implicit} and \textit{conflicted} variants in Section \ref{sec:human-baselines}. For the \textit{explicit} variant, we verify the questions are more straightforward than the \textit{implicit} set by checking that even smaller models can answer them. We find that all models 8B or larger that we test achieve $\geq$93\% multiple choice accuracy on the \textit{explicit} variant. 

\subsection{Human baselines}
\label{sec:human-baselines}
To measure human performance on the vignettes, we recruit four computer science PhD students\footnote{The annotators have no particular expertise in psychology or narrative, so we treat them as non-expert readers; three of the four grew up or spent substantial time in non-English-speaking countries, and individual cultural and linguistic backgrounds may shape how people read some narratives.}, who each answer 20 non-overlapping multiple choice questions from the \textit{implicit} and \textit{conflicted} variants, balanced across the four question types (interface in Appendix Figure~\ref{fig:interface}; results in Table \ref{tab:ToMStories} and Figure \ref{fig:conflicted-bars}). Annotators avoid the distractor answers on more than 80\% of the questions for both subsets, indicating that the vignettes are interpretable to human readers. Perfect accuracy is not expected, since the vignettes are synthetic and distractors can be close to the correct answer by chance (e.g., \textit{suspicion} and \textit{wariness}). For inter-annotator agreement, an author of this paper additionally annotates 40 of these vignettes, producing substantial agreement (Cohen's $\kappa = 0.73$; 74\% raw agreement). The author's ratings are excluded from human baselines in Table \ref{tab:ToMStories} as the author has particular expertise in the task. For the open-ended task, more than a month after completing the multiple choice task, two of the same annotators each answer \textasciitilde40 new questions (78 responses in total) that they had not seen in multiple choice form.   

\section{Methods}
We first benchmark model performance on the \textit{implicit} variant of ToMStories, and then explore how models are interpreting these hidden psychological aspects of characters through further experiments. 
\subsection{Benchmarking}
We test whether models can identify the correct answer from the four answer options (multiple choice scoring), as well as whether they can generate a reasonably close answer when not given answer options (open-ended scoring), which is substantially more challenging \cite{rottger-etal-2024-political} (full implementation details reported in Appendix \ref{sec:parsing}). 

\paragraph{Multiple choice scoring} For multiple choice questions, we compute accuracy based on whether a model selects the correct answer choice.

\paragraph{Open-ended scoring}
For the open-ended questions, model responses are free-form text rather than 
constrained to a multiple choice selection. We score these responses using two complementary 
methods: EmbSim and LLMScore. We embed the model's response and each of the four multiple choice answer options with \texttt{all-MiniLM-L6-v2}\footnote{\url{https://huggingface.co/sentence-transformers/all-MiniLM-L6-v2}} and compute two measures. Open-ended \textbf{accuracy} counts a response as correct if its embedding is closest, by cosine similarity, to the correct option. \textbf{EmbSim} is the cosine similarity between the response and the correct option, which captures how close a response is even when a distractor is closer. 
For LLM-based scoring (\textbf{LLMScore}), we use gpt-4o-mini as a judge to assign a Likert score from 0 to 4, indicating how similar the model's response $r$ is to the correct answer.

\paragraph{Models} We test a variety of open-source models from 1B to 72B parameters across the Llama, Qwen, Gemma, Phi, and DeepSeek model families (full list shown in Table \ref{tab:ToMStories}). See Appendix \ref{sec:hardware} for hardware details. 

\subsection{Generating indicator phrases}
\label{sec:phrase-methods}

To study whether models have a shared understanding of how to communicate and interpret hidden psychological states, we ask each model to \textit{list ten behavioral or physical cues a character might display if they are experiencing <psychological-state> but not talking about it openly} (full prompt in code repository). We sample 100 states used in ToMStories (25 per
question type) and prompt each of the 13 open-weight models and Claude Sonnet 4.6 to generate phrases for these states. We refer to these phrases as \textit{indicator phrases} and show examples in Appendix Table \ref{tab:phrase-examples}. 
\vspace{-0.2cm}
\paragraph{Overlapping indicator phrases.} We want to determine whether models have a shared understanding of what indicator phrases are associated with specific psychological states, and because models can describe the same character behavior in multiple ways, we compare indicator phrases by meaning rather than wording. 
After parsing and cleaning generated phrases (full details shown in Appendix~\ref{app:phrase-cleaning}), we
obtain 13{,}798 indicator phrases, which we embed with \texttt{all-MiniLM-L6-v2} and
cluster with average-linkage agglomerative clustering. We select 0.5 as the cosine distance
threshold for clustering as a reasonable threshold for this type of task with good performance on manual inspection. We expect this threshold is conservative as examples of clusters in Appendix Table \ref{tab:cluster-examples} demonstrate that phrases are coherently grouped (e.g., \textit{pauses excessively} and \textit{pausing for longer than usual} in the same cluster) but some clusters are still separate that could likely be combined (e.g., two different clusters focused on \textit{fidgeting}). We consider two phrases to \emph{overlap} in meaning if they fall in the same cluster, and we demonstrate in Appendix \ref{app:phrases-robustness} that our primary conclusions arising from this definition are robust to different choices of clustering threshold. 


\subsection{Embedding-based interpretability}
To probe model-internal representations of character psychology, we extract hidden-state 
activations from each layer of the evaluated transformer models. For a model with $L$ layers, 
let $h^{(\ell)}(x)$ denote the hidden state at layer $\ell \in \{1, \ldots, L\}$ 
at the last token position of input $x$, where the input includes the vignette, question, and answer options. For each narrative vignette $v$, we extract 
$\{h^{(\ell)}(v)\}_{\ell=1}^{L}$, yielding a sequence of representations that 
track the model's encoding of the vignette.

\paragraph{Linear probes}
To identify which layers contribute most to the model's ability to detect implicit psychological 
factors, we train linear probes on the per-layer embeddings described above. For each layer 
$\ell$, we fit a logistic regression classifier $f^{(\ell)}: \mathbb{R}^d \rightarrow \{1,2,3,4\}$ 
that maps the hidden state $h^{(\ell)}(v)$ to a predicted answer index:
\begin{equation*}
    f^{(\ell)}(h) = \underset{k \in \{1,2,3,4\}}{\arg\max} \; \mathbf{W}^{(\ell)} h + \mathbf{b}^{(\ell)}
\end{equation*}
where $\mathbf{W}^{(\ell)} \in \mathbb{R}^{4 \times d}$ and $\mathbf{b}^{(\ell)} \in 
\mathbb{R}^4$ are learned parameters and $d$ is the hidden state dimension. Probes are trained using cross-entropy loss with 
L2 regularization on the train split. 
We report layer-wise accuracy on the validation split. For each model, we select $\ell^*$ for further analysis as the layer at which the linear probe accuracy starts to plateau (equations in Appendix \ref{app:layer-selection}).

\paragraph{Steering vectors} To extract steering vectors, we first compute centroids 
for each question type $q \in \{\text{\textit{wound}, \textit{emotion}, \textit{fear}, 
\textit{motivation}}\}$ and for each variant $c \in \{\text{\textit{implicit}, \textit{explicit}}\}$:
\begin{equation*}
    \mu^{(\ell*)}_{q,c} = \frac{1}{|S_{q,c}|} \sum_{v \in S_{q,c}} h^{(\ell^*)}(v)
\end{equation*}
where $S_{q,c}$ is the set of vignettes of type $q$ under variant $c$. Following the mean-difference approach of \citet{sofroniew2026emotion},
we define the steering vector at layer $\ell*$ as
\begin{equation*}
    \mathbf{s}^{(\ell^*)} = \mu^{(\ell^*)}_{\textit{implicit}} - \mu^{(\ell^*)}_{\textit{explicit}}
\end{equation*}
where $\mu^{(\ell^*)}_{\textit{implicit}}$ and $\mu^{(\ell^*)}_{\textit{explicit}}$ are the implicit and
explicit centroids, pooled across all question types. This vector captures how the model's
representations of implicit and explicit vignettes differ. Steering generation along it lets us
interpret what that difference encodes. To steer, we add the unit-normalized vector
$\hat{\mathbf{s}}^{(\ell^*)} = \mathbf{s}^{(\ell^*)} / \lVert \mathbf{s}^{(\ell^*)} \rVert$, scaled by a
coefficient $\alpha$, to the residual stream at layer $\ell^*$ at every position during generation:
\begin{equation*}
    \tilde{h}^{(\ell^*)}(x) = h^{(\ell^*)}(x) + \alpha \, \hat{\mathbf{s}}^{(\ell^*)}
\end{equation*}
where $\alpha > 0$ steers toward more implicit expression and $\alpha < 0$ toward more explicit
expression. For each model, we tune $\alpha$ using gpt-4o-mini as a judge for coherence (full details in Appendix \ref{sec:steering-tuning}).

\section{Results}
We assess model performance on ToMStories, and then explore how models are interpreting implicit meaning in the vignettes.
\subsection{Benchmarking open models}

\begin{table*}[ht]
\centering\scriptsize
\setlength{\tabcolsep}{3.6pt}
\begin{tabular}{@{}l r ccccc ccccc cc@{}}
\toprule
 & & \multicolumn{5}{c}{\textbf{multiple choice accuracy (\%)}} & \multicolumn{5}{c}{\textbf{Open-ended accuracy (\%)}} & \multicolumn{2}{c}{\textbf{Open-ended score}} \\
\cmidrule(lr){3-7}\cmidrule(lr){8-12}\cmidrule(l){13-14}
Model & Size & Average & Wound & Emotion & Fear & Motivation & Average & Wound & Emotion & Fear & Motivation & EmbSim & LLMScore \\
\midrule
Llama-3.2-1B & 1B & 2.0 & 3.0 & 1.8 & 1.4 & 1.8 & \cellcolor{implicit!3}37.9 & 32.5 & \cellcolor{implicit!2}37.7 & \cellcolor{implicit!4}39.9 & \cellcolor{implicit!6}41.4 & 0.20 & \cellcolor{implicit!10}1.21 \\
Llama-3.2-3B & 3B & \cellcolor{implicit!17}53.9 & 34.1 & \cellcolor{implicit!29}67.5 & \cellcolor{implicit!18}55.5 & \cellcolor{implicit!20}58.2 & \cellcolor{implicit!10}46.4 & \cellcolor{implicit!3}37.9 & \cellcolor{implicit!15}51.6 & \cellcolor{implicit!10}46.5 & \cellcolor{implicit!13}49.4 & \cellcolor{implicit!22}0.29 & \cellcolor{implicit!12}1.23 \\
Llama-3.1-8B & 8B & \cellcolor{implicit!38}78.2 & \cellcolor{implicit!32}71.0 & \cellcolor{implicit!42}82.5 & \cellcolor{implicit!38}78.8 & \cellcolor{implicit!40}80.6 & \cellcolor{implicit!14}51.1 & \cellcolor{implicit!9}45.2 & \cellcolor{implicit!18}56.0 & \cellcolor{implicit!11}47.9 & \cellcolor{implicit!18}55.0 & \cellcolor{implicit!37}0.35 & \cellcolor{implicit!33}1.66 \\
Llama-3.1-70B & 70B & \cellcolor{implicit!47}\textbf{88.4} & \cellcolor{implicit!44}\textbf{85.6} & \cellcolor{implicit!46}87.5 & \cellcolor{implicit!48}\textbf{90.2} & \cellcolor{implicit!48}\textbf{90.2} & \cellcolor{implicit!18}55.6 & \cellcolor{implicit!17}54.4 & \cellcolor{implicit!25}63.7 & \cellcolor{implicit!14}51.3 & \cellcolor{implicit!16}53.0 & \cellcolor{implicit!45}0.38 & \cellcolor{implicit!45}1.90 \\
Llama-3.3-70B & 70B & \cellcolor{implicit!43}84.3 & \cellcolor{implicit!38}78.1 & \cellcolor{implicit!44}85.5 & \cellcolor{implicit!47}88.6 & \cellcolor{implicit!44}85.0 & \cellcolor{implicit!19}56.5 & \cellcolor{implicit!18}55.2 & \cellcolor{implicit!25}63.7 & \cellcolor{implicit!18}55.5 & \cellcolor{implicit!15}51.8 & \cellcolor{implicit!48}\textbf{0.39} & \cellcolor{implicit!45}1.90 \\
\addlinespace[2pt]
Qwen-2.5-7B & 7B & \cellcolor{implicit!33}73.1 & \cellcolor{implicit!22}60.0 & \cellcolor{implicit!39}79.8 & \cellcolor{implicit!35}74.9 & \cellcolor{implicit!37}77.4 & \cellcolor{implicit!16}52.7 & \cellcolor{implicit!12}48.3 & \cellcolor{implicit!21}58.7 & \cellcolor{implicit!12}49.1 & \cellcolor{implicit!17}54.8 & \cellcolor{implicit!32}0.33 & \cellcolor{implicit!31}1.63 \\
Qwen-2.5-72B & 72B & \cellcolor{implicit!46}87.2 & \cellcolor{implicit!42}82.8 & \cellcolor{implicit!47}\textbf{89.1} & \cellcolor{implicit!46}88.0 & \cellcolor{implicit!47}89.0 & \cellcolor{implicit!21}58.6 & \cellcolor{implicit!18}55.4 & \cellcolor{implicit!29}\textbf{67.7} & \cellcolor{implicit!15}52.5 & \cellcolor{implicit!21}\textbf{58.8} & \cellcolor{implicit!45}0.38 & \cellcolor{implicit!44}1.89 \\
\addlinespace[2pt]
Gemma-2-9B & 9B & \cellcolor{implicit!41}81.4 & \cellcolor{implicit!32}71.6 & \cellcolor{implicit!45}86.3 & \cellcolor{implicit!43}83.8 & \cellcolor{implicit!43}84.0 & \cellcolor{implicit!17}54.8 & \cellcolor{implicit!11}47.1 & \cellcolor{implicit!25}63.7 & \cellcolor{implicit!16}52.7 & \cellcolor{implicit!18}55.6 & \cellcolor{implicit!42}0.37 & \cellcolor{implicit!38}1.76 \\
Gemma-2-27B & 27B & \cellcolor{implicit!44}85.2 & \cellcolor{implicit!37}77.1 & \cellcolor{implicit!47}88.9 & \cellcolor{implicit!45}86.4 & \cellcolor{implicit!47}88.4 & \cellcolor{implicit!19}57.0 & \cellcolor{implicit!14}50.5 & \cellcolor{implicit!25}63.1 & \cellcolor{implicit!19}56.5 & \cellcolor{implicit!20}58.0 & \cellcolor{implicit!48}\textbf{0.39} & \cellcolor{implicit!45}1.91 \\
Mistral-7B & 7B & \cellcolor{implicit!20}57.8 & \cellcolor{implicit!10}46.7 & \cellcolor{implicit!32}71.0 & \cellcolor{implicit!24}62.9 & \cellcolor{implicit!14}50.8 & \cellcolor{implicit!16}53.7 & \cellcolor{implicit!14}51.5 & \cellcolor{implicit!22}60.5 & \cellcolor{implicit!13}49.3 & \cellcolor{implicit!16}53.4 & \cellcolor{implicit!22}0.29 & \cellcolor{implicit!49}\textbf{1.98} \\
Phi-4 & 14B & \cellcolor{implicit!39}79.1 & \cellcolor{implicit!27}66.1 & \cellcolor{implicit!43}83.7 & \cellcolor{implicit!43}83.6 & \cellcolor{implicit!42}82.8 & \cellcolor{implicit!19}56.4 & \cellcolor{implicit!12}49.1 & \cellcolor{implicit!25}63.1 & \cellcolor{implicit!17}54.5 & \cellcolor{implicit!21}\textbf{58.8} & \cellcolor{implicit!35}0.34 & \cellcolor{implicit!45}1.90 \\
\addlinespace[2pt]
R1-Distill-Qwen-7B & 7B & \cellcolor{implicit!14}51.0 & \cellcolor{implicit!2}37.1 & \cellcolor{implicit!23}61.1 & \cellcolor{implicit!13}49.3 & \cellcolor{implicit!19}56.2 & \cellcolor{implicit!4}39.6 & \cellcolor{implicit!1}36.3 & \cellcolor{implicit!4}40.1 & \cellcolor{implicit!5}41.1 & \cellcolor{implicit!5}41.0 & \cellcolor{implicit!15}0.26 & \cellcolor{implicit!4}1.08 \\
R1-Distill-Llama-8B & 8B & \cellcolor{implicit!34}74.3 & \cellcolor{implicit!24}62.9 & \cellcolor{implicit!41}81.7 & \cellcolor{implicit!34}73.9 & \cellcolor{implicit!38}78.6 & \cellcolor{implicit!11}48.1 & \cellcolor{implicit!7}42.6 & \cellcolor{implicit!16}53.2 & \cellcolor{implicit!10}46.5 & \cellcolor{implicit!13}50.0 & \cellcolor{implicit!27}0.31 & \cellcolor{implicit!37}1.74 \\
\midrule
GPT-4o-mini & -- & \cellcolor{implicit!42}82.9 & \cellcolor{implicit!35}74.6 & \cellcolor{implicit!44}85.5 & \cellcolor{implicit!44}85.6 & \cellcolor{implicit!45}86.0 & \cellcolor{implicit!15}52.6 & \cellcolor{implicit!9}45.8 & \cellcolor{implicit!24}62.7 & \cellcolor{implicit!11}47.7 & \cellcolor{implicit!17}54.0 & \cellcolor{implicit!40}0.36 & \cellcolor{implicit!24}1.48 \\
\midrule
\textit{Human} &  & \cellcolor{implicit!40}80.3 & \cellcolor{implicit!39}78.9 & \cellcolor{implicit!33}72.2 & \cellcolor{implicit!44}85.0 & \cellcolor{implicit!43}84.2 & \cellcolor{implicit!26}\textbf{64.2} & \cellcolor{implicit!29}\textbf{67.8} & \cellcolor{implicit!27}66.1 & \cellcolor{implicit!25}\textbf{64.0} & \cellcolor{implicit!21}\textbf{58.8} & \cellcolor{implicit!42}0.37 & \cellcolor{implicit!38}1.75 \\
\bottomrule
\end{tabular}
\caption{Performance on the \textit{implicit} ToMStories. Most Llama-3.2-1B outputs do not follow the answer format.}
\label{tab:ToMStories}
\end{table*}

In Table \ref{tab:ToMStories}, we benchmark open models of different sizes and families on both the multiple choice and open-ended questions of the \textit{implicit} vignettes. We include GPT-4o-mini as a reference of a closed foundation model to contextualize scores. We see that \textbf{on the multiple choice questions, the largest models can match human-level performance}, with Llama-3.1-70B achieving the best overall performance. Most models have the strongest performance on emotion questions. Interestingly, the DeepSeek reasoning-native variants of Qwen-2.5-7B and Llama-3.1-8B have lower scores across the board than the baseline models, suggesting that reasoning tokens may not be helpful for these tasks and models learn to represent subtext during pretraining. Given that performance plateaus for both humans and GPT-4o-mini, we expect the top scores represent roughly the saturation point for this data due to some answers being highly similar to distractors by chance (e.g., \textit{giving up a child} and \textit{navigating a changing family situation}). 

Accuracy on the open-ended questions is lower than humans for all models, with Qwen-2.5-72B performing the best. These results show that \textbf{humans are better than the models on the open-ended task} (64\% accuracy vs. 59\% for the best model), which remains true for GPT-4o-mini (53\% accuracy). Humans and models are similar on the open-ended emotion and motivation vignettes, but humans perform roughly 10\% better than models on wound and fear questions.

\begin{figure}[t]
  \centering
  \includegraphics[width=\columnwidth]{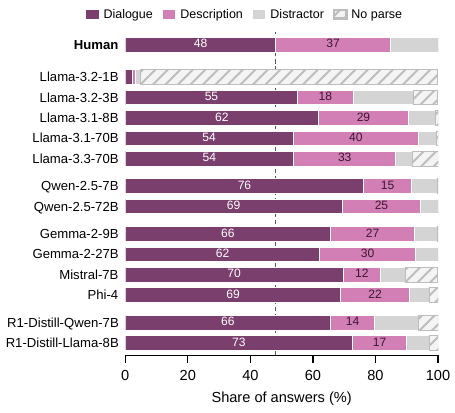}
  \caption{Answer selections on \textit{conflicted} ToMStories.}
  \label{fig:conflicted-bars}
\end{figure}

We next report results on the \textit{conflicted} variant of ToMStories in Figure \ref{fig:conflicted-bars}. In this case, there is not one correct answer, but an answer implied more by the description in the vignette and an answer implied more by the dialogue between characters, as well as two randomly selected distractors. We observe that models have strong performance at avoiding distractor answers (combined width of the pink bars), but that \textbf{models tend to choose more of the dialogue-aligned answers than the human would} (dark pink bars). The two Llama 70B models align most closely to the human breakdown, demonstrating a more even split between dialogue-aligned and description-aligned answers with a slight dialogue bias.

Given that these measures of interpretive ability are conducted with synthetic data, we have to be cautious in what conclusions we draw from them and we examine possible explanations for the strong model performance on \textit{implicit} vignettes in the subsequent sections.

\subsection{Evidence of shared indicator phrases}

\begin{figure*}[t]
  \centering
  \includegraphics[width=\textwidth]{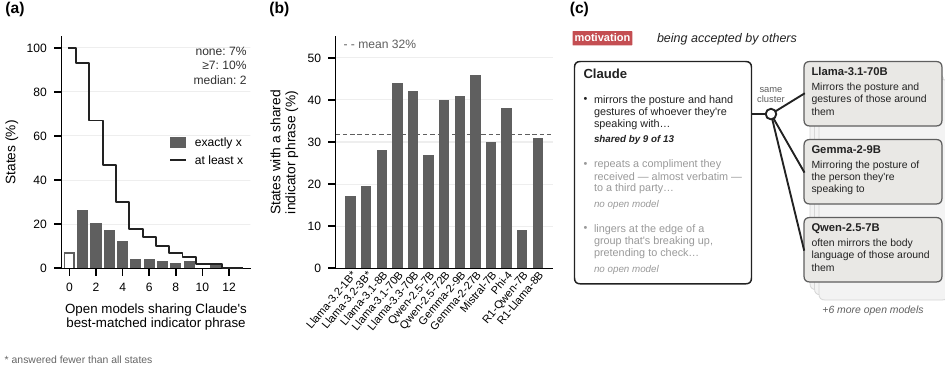}
\caption{Overlap between indicator phrases across models. \textbf{(a)} For each state, the number of open models with at least one phrase overlapping with Claude's. \textbf{(b)} Per model, the percentage of states with at least one phrase in the same cluster as one of Claude's. \textbf{(c)} Example for the
motivation state \textit{being accepted by others}. Claude's highest overlapping phrase overlaps with 9 of the 13 open models (which is a count for nine in panel A).}
  \label{fig:phrase-overlap}
\end{figure*}


We test whether there may be a shared set of indicator phrases models use to communicate and interpret hidden psychological states by evaluating overlap between open model indicator phrases and those of the synthetic data generator, Claude. In Figure \ref{fig:phrase-overlap}, we present this analysis. In Panel C, we show an example of a Claude-generated indicator phrase overlapping with other models' indicator phrases on a particular psychological state. In Panel A, we consider the most-overlapping phrase that Claude generated for each state, and count the number of other models that also generated that phrase. We see that Claude phrases have no overlap with other models for only 7\% of the states and for 10\% of states, Claude suggests a phrase that at least seven other models also suggest. In Panel B, we show for what percent of psychological states an individual model shares an indicator phrase with Claude. The strongest models have at least one overlapping phrase with Claude on over 40\% of psychological states. These numbers collectively present strong evidence for \textbf{a shared set of indicator phrases models associate with human psychological states}. Since a narrative likely includes multiple phrases, models do not need to overlap in all of their suggested phrases to receive meaningful interpretive signal. 


Indicator phrases capture how models generally conceptualize human states but may be different from what specifically appears in the generated vignettes, so in the next section we examine the connection between these phrases and model performance on ToMStories.

\subsection{Sharing indicator phrases with the data generator predicts model performance}

\begin{table}[t]
\centering
\scriptsize
\setlength{\tabcolsep}{3.5pt}
\begin{tabular}{lcccccc}
\toprule
 & \multicolumn{2}{c}{Pearson $r$} & \multicolumn{2}{c}{Regression $\beta$} & \multicolumn{2}{c}{\% of $R^2$} \\
\cmidrule(lr){2-3}\cmidrule(lr){4-5}\cmidrule(lr){6-7}
Predictor & MC & OE & MC & OE & MC & OE \\
\midrule
Model size (log params)   & .77$^{**}$ & .67$^{*}$  & .14 & .14       & 26 & 20 \\
Description preference    & .80$^{**}$ & .57        & .25 & $-$.29    & 29 & 14 \\
Phrase overlap w/ Claude  & \textbf{.89}$^{**}$ & \textbf{.92}$^{**}$ & \textbf{.60}$^{*}$ & \textbf{1.03}$^{**}$ & \textbf{45} & \textbf{66} \\
\midrule
Full model $R^2$          & \multicolumn{4}{c}{} & .85 & .88 \\
\bottomrule
\end{tabular}
\caption{A comparison of correlation between predictors and \textit{implicit} ToMStories accuracy, a multiple regression model fit to all three predictors and a dominance analysis of the general dominance of each of these predictors in this model. We exclude Llama-3.2-1B numbers from this analysis as the model failed to answer most questions. $^{*}p<.05$, $^{**}p<.01$}
\label{tab:predictors}
\end{table}

In the previous sections, we uncover several aspects of models that may contribute to their overall performance on \textit{implicit} ToMStories: model size, degree of description preference in interpretation, and overlap in indicator phrases with Claude (the vignette generator). We consider the log of a model's parameter count, the percent of description-aligned answers chosen on the \textit{conflicted} vignettes, and the overlap percents shown in Panel B of Figure \ref{fig:phrase-overlap} as proxies for these factors. In Table \ref{tab:predictors}, we report the Pearson's correlation of each of these predictors with \textit{implicit} accuracy (and plot these relationships in Appendix Figure \ref{fig:three-predictors}). We observe significant correlation between all three factors and both multiple choice and open-ended accuracy (except for description preference and open-ended accuracy). Indicator phrase overlap with Claude has the strongest correlation with model performance ($r = 0.89$, $p<.01$ for multiple choice; $r = 0.92$, $p<.01$ for open-ended). 

Because these three predictors are themselves correlated across models ($r$ = .73–.75), we estimate their relative contributions jointly with a multiple regression with standardized variables and dominance analysis for general dominance (also reported in Table \ref{tab:predictors}). Indicator phrase overlap is the only significant predictor when all three are included (multiple choice $\beta$ = 0.60, $p$ = .023; open-ended $\beta$ = 1.03, $p$ = .001). Indicator phrase overlap also accounts for the largest share of explained variance (45\% and 66\% of $R^2$, vs. $\leq$29\% for other predictors). For open-ended questions, this finding is robust to different clustering thresholds for phrase overlap (detailed in Appendix \ref{app:phrases-robustness}). These results suggest that \textbf{not only do models learn a shared set of indicator phrases, but the degree to which their indicator phrases overlap with the synthetic data generator is the strongest predictor of interpretive performance}. 
Shared indicator phrases may let models ``shortcut’’ interpretation of synthetic data, or they may reflect general social reasoning ability. Testing whether Claude’s indicator phrases match how humans describe behavior would distinguish these accounts. Overall, these results help us understand how models perform interpretation, while indicating that synthetic data likely simplifies the problem. We explore another simplification models may use next. 

\subsection{Implicit means less emotional to models}
\label{sec:embeddings}
\begin{figure}[t]
  \centering
  \includegraphics[width=.9\columnwidth]{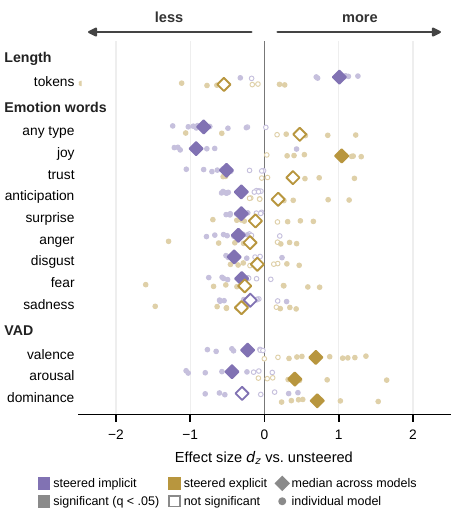}
  \caption{Effect of implicit (purple) and explicit (gold) steering relative to unsteered generations from the same prompts. Diamonds show the median paired effect size $d_z$	across models, filled where significant across models (Wilcoxon, BH-corrected). Dots show individual models, filled where significant ($q<.05$).}
  \label{fig:lexicon-baseline}
\end{figure}

Linear probes trained to predict the correct answer gain accuracy at
similar layers for explicit and implicit vignettes, but reach higher accuracy on explicit ones (shown in Appendix Section \ref{sec:linearprobes}). This suggests that models process both kinds of vignette
similarly, but explicit vignettes carry more decodable ``signal.'' A possible explanation for this signal is presented by considering the differences in model representations for implicit vs. explicit questions, extracted as a steering vector (visualized in Appendix Section \ref{app:embeddings}).

To characterize what this direction in activation space captures, we generate vignettes steered toward the explicit and implicit directions, as well as unsteered vignettes, using the same random sample of 100 ToMStories seeds for each. Qualitatively, implicit steering
produces longer, more ornate prose, whereas explicit steering produces shorter, more direct
text (examples in Appendix Table~\ref{tab:steering-examples}). To quantify the change in emotional
language, we score each vignette with the NRC Emotion Intensity Lexicon \citep{LREC18-AIL}
(the rate of words associated with each of eight emotions) and the NRC VAD
Lexicon \citep{mohammad2025nrcvadlexiconv2} (the mean valence, arousal, and dominance of its
words). For each model and measure, we compare steered and unsteered vignettes with a paired
$t$-test, and we test for a consistent effect across models with a Wilcoxon signed-rank test on
the per-model mean differences, correcting for multiple measures with Benjamini--Hochberg. We exclude Llama-3.2-1B and the two R1 distilled models,
whose steered outputs frequently degenerate.

In Figure~\ref{fig:lexicon-baseline}, we see that \textbf{implicit steering consistently suppresses emotional language.} Relative to unsteered
text, implicit vignettes contain fewer emotion words for seven of the eight emotions (all but
sadness) and have lower valence and arousal. 
This pattern is
consistent with \citet{sofroniew2026emotion}, which found a steering vector for masking emotion through a related process. While muting emotion is one method of implicit communication, using emotional signal as a primary axis of modulation between implicit and explicit scenarios is a simplification of human expression. In reality, implicit psychological expression should and can still be quite emotional. 

\section{Conclusion}
We introduce ToMStories as a resource for the community and use it to show how models interpret and process character psychology in narrative. We see strong evidence that models 
learn shared indicator phrases that are associated with human psychological states and play a role in interpreting subtext. We also see that suppressing emotional language is part of how models communicate psychology implicitly, and that models rely more on dialogue than description to access character internal states. Collectively, these results show how working in the controlled setting of synthetic data can help us formulate and test hypotheses about how models perform interpretation, while also serving as a caution against taking model performance on synthetic human psychology as representative of performance on real human data. Future work should test whether indicator phrases causally shape model interpretation, and whether
humans share the indicator phrases models use.  


\section*{Limitations}

\paragraph{Synthetic data from a single generator.}
All ToMStories vignettes are written by one model, Claude Sonnet 4.6. The benchmark therefore
reflects one model's conventions for how inner states surface in behavior and speech, and our
results show that models whose indicator phrases resemble Claude's tend to score higher. As discussed and elaborated on in our results, accuracy on
ToMStories may thus partly measure alignment with the generator's conventions rather than
interpretive ability in general, and it may overestimate how well models interpret the more varied
and less conventional indicators found in human-written narratives. Vignettes from other generators,
or written by humans, would be needed to separate these factors.

\paragraph{Cultural scope and human baselines.}
The states we test come from a thesaurus written for Western, English-language writers, so both
the states and the behaviors that signal them reflect one cultural framing of psychology. Our
human baselines come from a small number of annotators from varied cultural and language
backgrounds; they estimate human performance only roughly, and differences between the
thesaurus's framing and annotators' own conventions may contribute to human performance.

\paragraph{Representation and steering analyses.}
Our probing and steering analyses are limited to open-weight models, so we cannot examine the
generator's internal representations. Additionally, our layer-selection heuristic involves fixed choices of
smoothing and threshold, and steering quality is sensitive to layer selection.

\section*{Acknowledgments}
We are grateful to the Columbia University students who contributed annotations, and to Mohit Iyyer and Katherine Thai for their helpful feedback on early versions of this work. One of the authors,
Melanie Subbiah, holds an equity interest in OpenAI.


\bibliography{custom}
\clearpage
\appendix


\section{Layer-wise Analysis}
\label{sec:linearprobes}
\begin{figure*}[t]
    \centering


    \includegraphics[width=\linewidth]{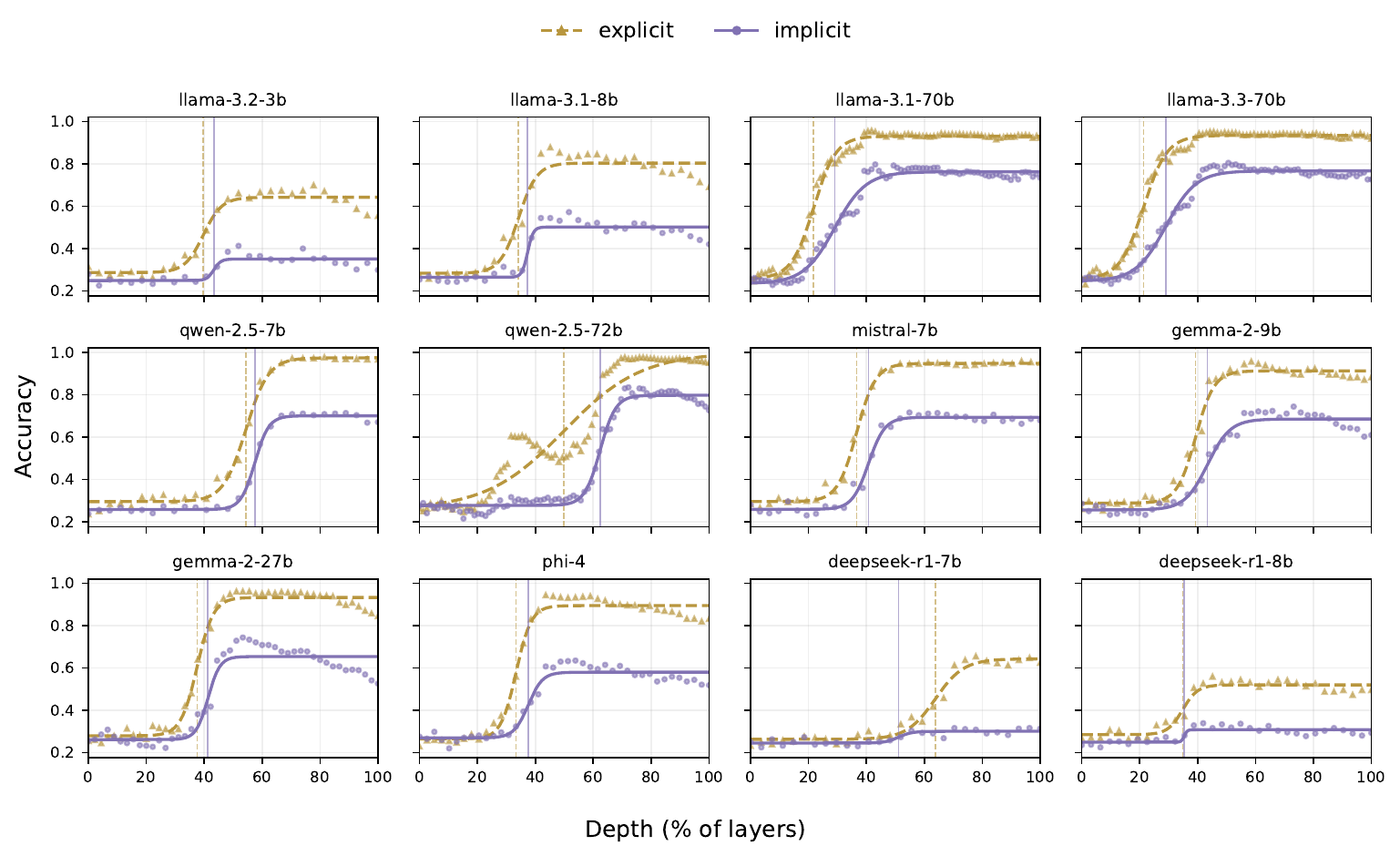}
    \caption{Classification accuracy for linear probes trained to predict the correct answer from the outputs of each layer. We report overall accuracy, as well as split across the \textit{implicit} and \textit{explicit} variants of ToMStories with sigmoids fit to the data and vertical lines indicating the medians of the sigmoid curves.}
    \label{fig:layer-accuracy}
\end{figure*}

We show some examples of layer-accuracy trajectories for linear probes in Figure \ref{fig:layer-accuracy}. For smaller models, there seems to be a jump in accuracy over about five layers just before the half-way point of the model. For Qwen 
models, this improvement comes after the half-way point, indicating that earlier layers perform more preprocessing, possibly because Qwen is not an English-base model. For the reasoning-native DeepSeek models, we see that the embedding-based accuracy underperforms the model's ultimate accuracy, so more of the performance is gained in the inference-time reasoning and how that updates internal representations. For larger models, like Llama-70B, we observe an improvement in accuracy across more layers in the model, demonstrating the more complex processing necessary to achieve higher ultimate accuracy.

In addition to overall accuracy, we split out classification accuracy based on the \textit{implicit} and \textit{explicit variants} of ToMStories in Figure \ref{fig:layer-accuracy}. Interestingly, we observe that \textbf{implicit vignettes are not classified later in the layers but almost directly track classification improvements in \textit{explicit} scenarios} (albeit with lower accuracy). 
This observation suggests that the models are processing both implicit and explicit scenarios using similar mechanisms, but for explicit scenarios there is stronger signal in the data for the answer.

\subsection{Layer selection}
\label{app:layer-selection}
The analyses in Section \ref{sec:embeddings} operate at a single
layer per model, which we select from the probe curves.
We select the \emph{onset} of the plateau in probe accuracy: the
shallowest layer at which the answer is already mostly linearly decodable. We first smooth the
held-out probe accuracies across layers with a 3-layer moving average (truncated at the first
and last layers). Let $a_{\max}$ be the peak of the smoothed curve and $a_{\min}$ the lowest
smoothed accuracy at or before that peak. We set a threshold 90\% of the way up this rise,
\begin{equation*}
\tau = a_{\min} + 0.9\,(a_{\max} - a_{\min}),
\end{equation*}
and select the earliest layer, at or before the peak, whose smoothed accuracy reaches $\tau$
and stays at or above it for the next layer as well (i.e., two consecutive layers). If no layer
meets this condition, we use the peak layer. Taking $a_{\min}$ only over layers up to the peak
keeps a drop in accuracy at late layers from lowering the threshold, and requiring two
consecutive layers keeps a single-layer spike from moving the selected layer shallower.

\section{Hardware Details}
\label{sec:hardware}
We run most models on one 40GB A100 GPU. For models larger than 25B parameters, we use two 40GB A100 GPUs and 4-bit quantization with double quantization using either AWQ or bitsandbytes. We use temperature 0 and allow most models to generate up to 400 tokens. For the reasoning-native DeepSeek models, we allow 1500 tokens.

\section{Embedding Analysis}
\label{app:embeddings}

\begin{figure*}[t]
  \centering
  \includegraphics[width=.9\textwidth]{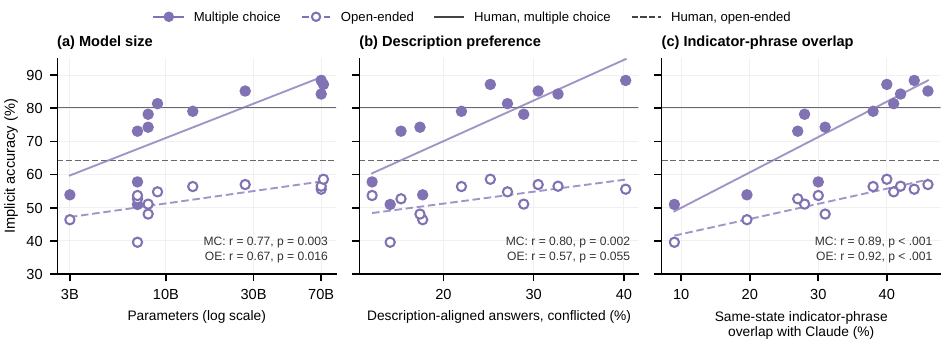}
  \caption{Implicit accuracy against three model properties, across 12 models (Llama-3.2-1B
  excluded due to many unparseable answers): \textbf{(a)} parameter count, \textbf{(b)} the share of description-aligned answers selected on \textit{conflicted} ToMStories, and \textbf{(c)} indicator phrase overlap with Claude (panel B in Figure \ref{fig:phrase-overlap}). Filled markers and solid lines show multiple choice; open markers and dashed lines show
  open-ended; horizontal lines mark human performance.}
  \label{fig:three-predictors}
\end{figure*}

\begin{table}[t]
\centering \small
\begin{tabular}{lrrrr}
\toprule
\textbf{Model} & $\ell^*$& $\overline{\cos(d_t,d_{t'})}$ & $\mathrm{CV}(\|d_t\|)$ \\
\midrule
  deepseek-r1-7b & 20 &  0.616 & 0.303 \\
  deepseek-r1-8b & 13 &  0.518 & 0.090 \\
  gemma-2-27b & 21  & 0.660 & 0.201 \\
  gemma-2-9b & 22  & 0.737 & 0.373 \\
  llama-3.1-70b & 31 &  0.656 & 0.238 \\
  llama-3.1-8b & 13 &  0.533 & 0.076 \\
  llama-3.2-1b & 10 &  0.512 & 0.149 \\
  llama-3.2-3b & 14 &  0.559 & 0.062 \\
  llama-3.3-70b & 31 &  0.663 & 0.158 \\
  mistral-7b & 15  & 0.788 & 0.315 \\
  phi-4 & 17  & 0.663 & 0.147 \\
  qwen-2.5-72b & 54 &  0.596 & 0.150 \\
  qwen-2.5-7b & 18  & 0.720 & 0.306 \\
\midrule
  Average & & 0.632 & 0.198 \\
\bottomrule
\end{tabular}
\caption{High-dimensional factorization statistics per model and layer ($l$) for steering vectors. All metrics are averaged across the four question types.  $\overline{\cos(d_t, d_{t'})}$: mean pairwise cosine between all steering vectors; CV$(\|d_t\|)$: coefficient of variation of steering vector magnitudes.} 
\label{tab:centroidstats}
\end{table}

\begin{figure*}[p]
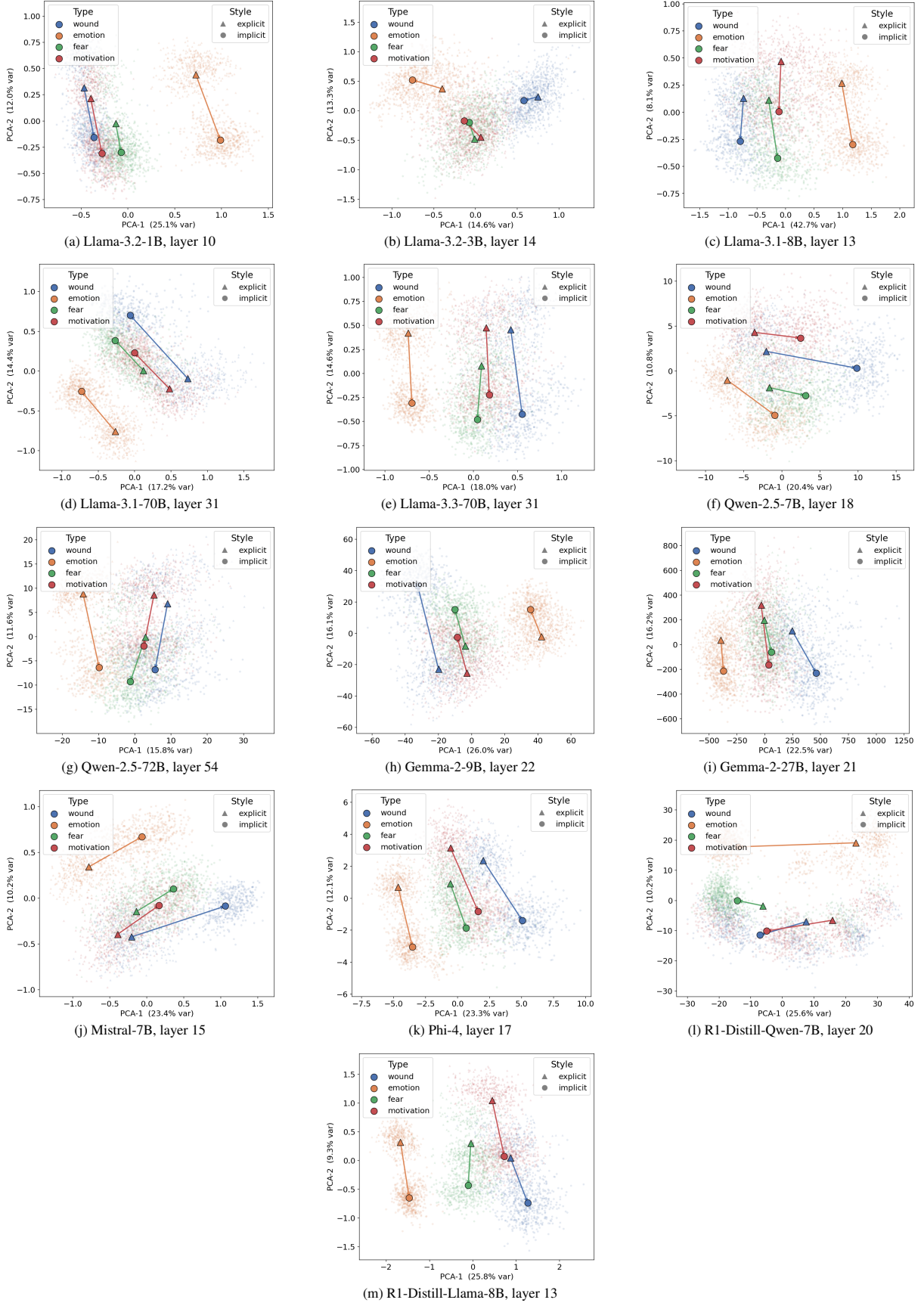

    \centering
    \captionsetup[subfigure]{font=scriptsize, skip=1pt}
    \newcommand{\pcapanel}[3]{%
        \begin{subfigure}[b]{0.30\textwidth}
            \centering
            \includegraphics[width=\textwidth]{figures/#1_layer#2_centroid_pca_pc12.png}
            \caption{#3, layer #2}
            \label{fig:pca-#1}
        \end{subfigure}}
    \pcapanel{llama-3.2-1b}{10}{Llama-3.2-1B}\hfill
    \pcapanel{llama-3.2-3b}{14}{Llama-3.2-3B}\hfill
    \pcapanel{llama-3.1-8b}{13}{Llama-3.1-8B}\\[4pt]
    \pcapanel{llama-3.1-70b}{31}{Llama-3.1-70B}\hfill
    \pcapanel{llama-3.3-70b}{31}{Llama-3.3-70B}\hfill
    \pcapanel{qwen-2.5-7b}{18}{Qwen-2.5-7B}\\[4pt]
    \pcapanel{qwen-2.5-72b}{54}{Qwen-2.5-72B}\hfill
    \pcapanel{gemma-2-9b}{22}{Gemma-2-9B}\hfill
    \pcapanel{gemma-2-27b}{21}{Gemma-2-27B}\\[4pt]
    \pcapanel{mistral-7b}{15}{Mistral-7B}\hfill
    \pcapanel{phi-4}{17}{Phi-4}\hfill
    \pcapanel{deepseek-r1-7b}{20}{R1-Distill-Qwen-7B}\\[4pt]
    \pcapanel{deepseek-r1-8b}{13}{R1-Distill-Llama-8B}
    \caption{Centroids of implicit and explicit vignettes for each question type, projected onto the
    first two principal components, at the layer selected by linear probe analysis.}
    \label{fig:centroids}
\end{figure*}

We plot cluster centroids in Figure \ref{fig:centroids} for vignette embeddings across variants and types. We additionally compute                                                                                
  two summary statistics directly in the high-dimensional activation space and                                                                                
  report them in Table~\ref{tab:centroidstats} to support the PCA plots.                                                                                                                                                                                                                                               
  Let $\mu(t, c) \in \mathbb{R}^d$ denote the centroid of all last-token                                                                                       
  activations at layer $\ell^*$ for question type $t \in \{\text{wound, emotion, fear,                                                                                           
  motivation}\}$ and variant $c \in \{\text{implicit, explicit}\}$.                                                                                              
  Define the per-type variant-shift vector $d_t = \mu(t, \text{implicit}) -                                                                                      
  \mu(t, \text{explicit})$ and the mean shift $\bar{v} = \frac{1}{4}\sum_t d_t$.                                                                               
  Under perfect additive factorization, all $d_t$ are identical (and equal to                                                                                  
  $\bar{v}$).                                                                                                                                                                                                                     
  \textbf{CV}$(\|d_t\|)$ is the coefficient of variation (standard deviation                                                                                   
  divided by mean) of the four shift magnitudes. A value near 0 indicates that                                                                                 
  the variant transformation is consistent in magnitude across types as well as                                                                                  
  direction.                                                                                                                                                   
  \textbf{Mean} $\cos(d_t, d_{t'})$ is the average pairwise cosine similarity                                                                                
  between all six pairs of shift vectors.


\begin{table*}[t]
\centering\scriptsize
\setlength{\tabcolsep}{4pt}
\renewcommand{\arraystretch}{1.2}
\begin{tabular}{@{}>{\RaggedRight\arraybackslash}p{0.11\textwidth}
                >{\RaggedRight\arraybackslash}p{0.275\textwidth}
                >{\RaggedRight\arraybackslash}p{0.275\textwidth}
                >{\RaggedRight\arraybackslash}p{0.275\textwidth}@{}}
\toprule
\multicolumn{4}{@{}l}{\textbf{Prompt:} write a short narrative. \textit{Setting:} nursery;
  \textit{relationship:} friends; \textit{emotion:} betrayed} \\
\midrule
\textbf{Model} & \textbf{Unsteered} & \cat{explicit}{\textbf{Steered toward explicit}} &
  \cat{implicit}{\textbf{Steered toward implicit}} \\
\midrule
Llama-3.1-8B &
Lily's eyes welled up with tears as she stared at the blocks her friend Emma had built into a
magnificent castle. Just yesterday, Emma had promised to let her add a tower to the castle, but now
it was gone, replaced by a bright pink flag that read ``Emma's Castle'' in bold letters. Lily felt a
pang of betrayal.\,\ldots \wc{247} &
Writing a short narrative based on this scenario is straightforward. Here is a short narrative based
on this scenario. A nursery is a place where young children learn and play. It is a place of joy and
innocence. However, when a person is feeling betrayed, the joy and innocence are replaced with
feelings of hurt and anger. Here is a short narrative based on this scenario.\,\ldots \wc{430} &
In the dimly lit nursery, Emma sat amidst the rows of cribs, her eyes fixed on the small, unmade bed
that seemed to be a miniature version of the one she shared with her best friend, Olivia. The two
girls had been inseparable since they were toddlers, and their friendship was a complex web of
whispered secrets and stolen glances.\,\ldots \wc{378} \\
\addlinespace[3pt]
Qwen-2.5-72B &
In the soft glow of the nursery, where the walls were painted with whimsical forest scenes and the
air was filled with the scent of lavender, two friends, Mia and Lily, sat among the toys and books.
\ldots\ Today, however, the atmosphere was different. Mia's eyes were red from crying,\,\ldots \wc{370} &
Lily and Max were best friends. They were playing with toys in the nursery. Lily was feeling very
betrayed by Max because he had told her a huge lie. She looked at Max with tears in her eyes and
said, ``I can't believe you lied to me!'' Max felt terrible for lying to his friend. He apologised
and promised to never do it again.\,\ldots \wc{73} &
In the quiet of the nursery, the air was filled with the scent of baby powder and the soft hum of a
lullaby from a distant room. The nursery was not just for the baby; it had become a sort of neutral
ground where the two friends, Mia and Alex, often met. They were not the type to openly discuss
their feelings,\,\ldots \wc{425} \\
\addlinespace[3pt]
Gemma-2-27B &
The rocking horse stood frozen mid-gallop, its painted eyes staring vacantly at the wall. Lily
traced a finger along its chipped mane, the wood rough beneath her touch. It used to be their horse,
hers and Maya's.\,\ldots\ But Maya wasn't here anymore. Not since she'd told everyone about Lily's
secret,\,\ldots \wc{232} &
The pastel walls of the nursery glowed under the warm light of the ceiling lamp. Lily bounced
excitedly, pointing at the crib. ``Isn't he cute?'' she giggled. I smiled back, my heart swelling
with pride. ``He's perfect,'' I said, gently rocking the crib. Lily clapped her hands. ``Let's sing
him a song!''\,\ldots \wc{332} &
The nursery was a tableau of faded pastels and forgotten dreams. Dust motes danced in the shafts of
sunlight filtering through the grimy window, illuminating the room's melancholic stillness. Elara,
perched on the window seat, was a study in contradictions. Her posture, with its slumped shoulders
and downcast gaze, suggested vulnerability,\,\ldots \wc{374} \\
\bottomrule
\end{tabular}
\caption{Examples of unsteered and steered generations for the same prompt from three models.}
\label{tab:steering-examples}
\end{table*}

\section{Steering Coefficient Tuning}
\label{sec:steering-tuning}

\begin{algorithm}[t]
\scriptsize
\caption{Steering coefficient tuning}
\label{alg:steer-tuning}
\begin{algorithmic}
\Require model, steering layer, unit steering vector $\hat{v}$, writing prompts
\Ensure a coefficient per direction, and one shared across directions

\State $scale \gets$ median residual-stream norm at the steering layer
\State $baseline \gets$ mean coherence of unsteered generations
\State $floor \gets baseline - 1$
  \Comment{tolerate a 1-point drop}

\ForAll{$direction \in \{+1, -1\}$}
  \ForAll{$\alpha \in \{0.1, 0.2, 0.3, 0.5, 0.75, 1, 1.5, 2\}$}
    \State steer by $\textit{direction}\cdot \alpha \cdot scale \cdot \hat{v}$ at every position
    \State $coherence[\alpha] \gets$ mean judge score over the prompts
  \EndFor
  \State $best \gets$ largest $\alpha$ with $coherence[\alpha]\geq$ \textit{floor}
  \State \textbf{repeat} 3 times:
  \State \quad test the midpoint between $best$ and the smallest failing $\alpha$
  \State \quad update $best$ if the midpoint coherence clears $floor$
  \State $coefficient[direction] \gets best \cdot scale$
\EndFor

\State $shared \gets \min$ of $best$ over the two directions
\State verify $shared$ clears $floor$ in both directions, else flag the model
\State \Return $coefficient[\cdot]$ and $shared \cdot scale$
\end{algorithmic}
\end{algorithm}

To normalize steering strength across models whose residual streams differ in scale, we parameterize it relative to each model's own activations. We measure $\rho$, the median residual-stream norm at the steering layer $\ell$ over the tokens of a set of probe prompts (excluding each sequence's first token, which is generally an outlier) and apply $\rho$ as a scaling factor during steering. We consider both steering directions, $+1$ for implicit and $-1$ for explicit. For each direction, we search a range of alphas for the largest value that does not degrade fluency on a set of 12 generic writing prompts. To compare steering directions, we adopt the smaller of the two per-direction magnitudes as a shared $\alpha^{\pm}$. This tuning process is described in more detailed pseudocode in Algorithm \ref{alg:steer-tuning}.

\section{Benchmark Scoring}
\label{sec:parsing}
For multiple choice scoring, we consider an answer correct if the model either places the number of the correct answer choice between <answer></answer> tags as the prompt instructs, or the number of the correct answer is the last number mentioned in the last 30 characters of the response.

For open-ended responses, responses that do not 
follow the specified output format are marked as unparseable and excluded from accuracy 
calculations.

For embedding-based accuracy, let $r$ denote the model's free-form response and let 
$\{a_1, a_2, a_3, a_4\}$ denote the four multiple choice options for a given vignette. We embed each string using a fixed encoder $\phi(\cdot)$, and report the cosine similarity 
between the model's response and each answer option as a continuous measure of response quality:
\begin{equation*}
    \textbf{EmbSim}(r, a) = \frac{\phi(r)^\top \phi(a)}{\|\phi(r)\| \|\phi(a)\|}
\end{equation*}
To obtain an accuracy figure 
comparable to the multiple choice variant, we assign the model the answer $\hat{a}$ with the 
highest similarity to the response:
\begin{equation*}
    \hat{a} = \underset{a_i}{\arg\max} \; \textbf{EmbSim}(r, a_i)
\end{equation*}
We use all-MiniLM-L6-v2\footnote{\url{https://huggingface.co/sentence-transformers/all-MiniLM-L6-v2}} as $\phi(\cdot)$. 

For LLM-based scoring, we use an LLM-as-a-judge protocol in which a judge model $\mathcal{J}$ 
assigns a Likert score from 0 to 4, indicating how similar the model's response $r$ is to the correct answer $a^*$ given the question $q$:
\begin{equation*}
    \textbf{LLMScore} = \mathcal{J}(r, a^*, q)
\end{equation*}
We use gpt-4o-mini as the judge model and the prompt is shown in our repository.

\section{Human Baseline Interface}  
\begin{figure*}[t]
    \centering
    \includegraphics[width=\linewidth]{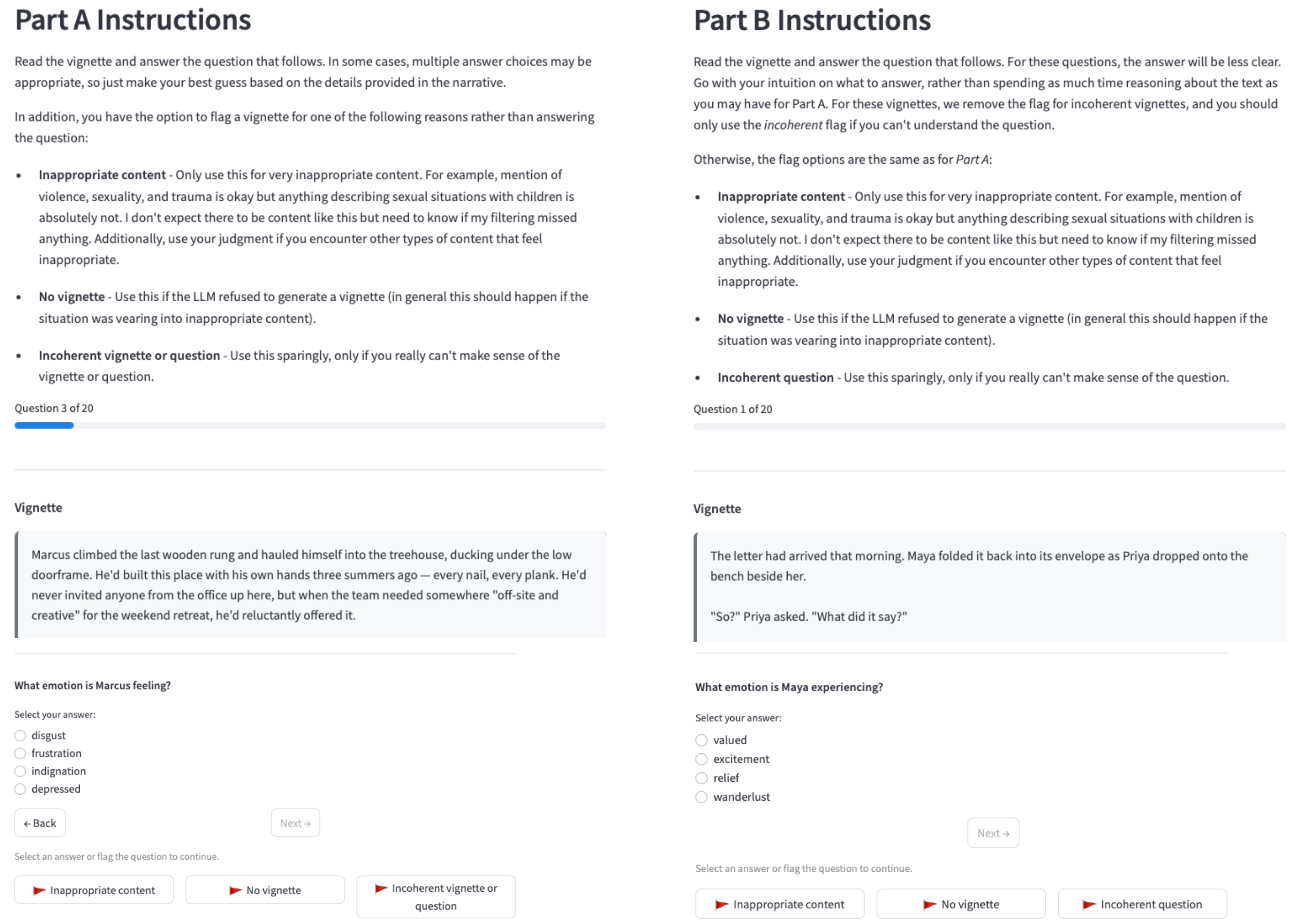}
    \caption{The two interfaces for collecting the human baselines. Part A is for the implicit vignettes, and part B is for the conflicted vignettes. We do not display the full vignette text here for brevity.}
    \label{fig:interface}
\end{figure*}

The interface for collecting the human baseline is shown in Figure \ref{fig:interface}. In addition to the ToMStories question, we ask the annotators to check for inappropriate content, model refusals, and incoherent vignettes or questions that cannot be answered. No vignettes are flagged as inappropriate or model refusals. Two vignettes/questions are flagged as incoherent in the set of 80 \textit{implicit} vignettes annotated by humans and none of the 80 \textit{conflicted} vignettes.

\label{sec:human_baseline}             


\begin{table}[ht]
\centering\scriptsize
\setlength{\tabcolsep}{2pt}
\begin{tabular}{@{}lcc>{\columncolor{neutral!10}}ccc@{}}
\toprule
 & \multicolumn{5}{c}{\textbf{Clustering threshold}} \\
\cmidrule(l){2-6}
Measure & 0.40 & 0.45 & \textbf{0.50} & 0.55 & 0.60 \\
\midrule
Clusters & 4,256 & 3,111 & 2,188 & 1,464 & 958 \\
\multicolumn{6}{@{}l}{\textit{Overlap (all phrases, any state)}} \\
\quad $\geq$1 other model (\%) & 79 & 87 & 92 & 96 & 98 \\
\quad $\geq$7 other models (\%) & 33 & 43 & 55 & 66 & 77 \\
\quad Median other models & 3 & 5 & 8 & 10 & 12 \\
\multicolumn{6}{@{}l}{\textit{Overlap and implicit acc.}} \\
\quad Pearson $r$ (overlap), MC & $0.84^{**}$ & $0.87^{**}$ & $0.89^{**}$ & $0.82^{**}$ & $0.74^{**}$ \\
\quad \phantom{Pearson $r$ (overlap),} OE & $0.86^{**}$ & $0.90^{**}$ & $0.92^{**}$ & $0.91^{**}$ & $0.92^{**}$ \\
\multicolumn{6}{@{}l}{\textit{Regression on MC acc.}} \\
\quad $\beta$: model size & $0.07$ & $-0.01$ & $0.14$ & $0.07$ & $0.12$ \\
\quad \phantom{$\beta$:} description pref. & $0.42$ & $0.37$ & $0.25$ & $0.47^{*}$ & $0.51$ \\
\quad \phantom{$\beta$:} overlap & $0.52^{*}$ & $0.61$ & $0.60^{*}$ & $0.53^{*}$ & $0.41$ \\
\quad \% of $R^2$: model size & 27 & 26 & 26 & 26 & 29 \\
\quad \phantom{\% of $R^2$:} description pref. & 34 & 32 & 29 & 34 & 39 \\
\quad \phantom{\% of $R^2$:} overlap & 40 & 42 & 45 & 39 & 32 \\
\multicolumn{6}{@{}l}{\textit{Regression on OE acc.}} \\
\quad $\beta$: model size & $0.07$ & $-0.18$ & $0.14$ & $0.06$ & $0.02$ \\
\quad \phantom{$\beta$:} description pref. & $0.01$ & $-0.11$ & $-0.29$ & $0.08$ & $0.16$ \\
\quad \phantom{$\beta$:} overlap & $0.80^{*}$ & $1.12^{**}$ & $1.03^{**}$ & $0.82^{**}$ & $0.83^{**}$ \\
\quad \% of $R^2$: model size & 24 & 22 & 20 & 21 & 21 \\
\quad \phantom{\% of $R^2$:} description pref. & 15 & 14 & 14 & 13 & 13 \\
\quad \phantom{\% of $R^2$:} overlap & 61 & 65 & 66 & 66 & 66 \\
\bottomrule
\end{tabular}
\caption{Robustness of the indicator-phrase clustering results to the clustering threshold; the shaded column (0.50) is used in the main text.}
\label{tab:threshold-robustness}
\end{table}

\section{Indicator phrases}
\subsection{Parsing and cleaning generated phrases}
\label{app:phrase-cleaning}
We request a bare JSON array of short strings from models but expect formatting failures so adopt the following parsing approach. We first remove any <think> reasoning trace, then look for a fenced code block containing an array or a span between square braces and parse that as JSON. If parsing fails, we fall back to treating the response as a plain or numbered list, one cue per line, removing any numbered item start. We split long answers that are several phrases concatenated (truncated lists, or 6+ comma-separated fragments) and strip any residual JSON and Markdown debris. Lastly, we discard phrases that are shorter than four characters, longer than 300 characters, only punctuation, beginning with conversational preamble from the model, or duplicating another phrase in the model's list. Before clustering, we reduce each phrase to its core behavior by removing any leading subordinate clause, truncating at the first hard delimiter or subordinating conjunction, and dropping any leading pronoun subject. For example, "Avoids eye contact when her mother calls" and "She avoids eye contact, shoulders tensing" are both simplified to "avoids eye contact". These final phrases are 7.3 words long on average.

\subsection{Clustering threshold robustness}
\label{app:phrases-robustness}

Indicator phrase overlap depends on the threshold used in agglomerative clustering, so we repeat our
overlap analyses with clustering thresholds from 0.40 to 0.60 (results in Table~\ref{tab:threshold-robustness}). We see that higher thresholds merge more phrases, resulting in higher overlap numbers as expected. The substantive conclusions hold across thresholds: (i) same-state overlap correlates with implicit accuracy across models
($r = .74$--$.89$ for multiple choice and $.86$--$.92$ for open-ended; all $p < .01$), and (ii) controlling for model size and description preference, overlap remains a significant predictor of implicit accuracy at every threshold for open-ended and most thresholds for multiple choice.

\section{AI Use}
We used Claude Code to support with the implementation and visualization of experiments. All ideas came from us and we provided Claude Code with explicit instructions for what to implement or modify. We ran all of the experiments ourselves and verified the code. In writing the paper, we used Claude and ChatGPT as assistants with literature review, summarizing claims, and re-wording complex sentences.

\begin{table*}[t]
\centering\small
\setlength{\tabcolsep}{4pt}
\renewcommand{\arraystretch}{1.2}
\begin{tabular}{@{}>{\raggedright\arraybackslash}p{0.17\textwidth}r >{\raggedright\arraybackslash}p{0.695\textwidth}@{}}
\toprule
\textbf{Cluster label} & \textbf{\textit{n}}  & \textbf{Example indicator phrases} \\
\midrule
Avoids eye contact & 722  & avoiding eye contact that lingers slightly; avoiding eye contact with mirrors; Avoiding eye contact to hide nervousness or guilt about feelings; Maintaining a closed posture with crossed arms and avoiding eye contact; avoids eye contact with certain individuals or groups; avoids eye contact by looking around the room nervously; avoiding eye contact with acquaintances; Slouching in financial discussions to avoid eye contact; avoids eye contact and fidgets; avoided eye contact with others, lost in their own thoughts \\
\addlinespace[2pt]
Types of pauses & 290  & pauses to collect own thoughts; Pauses to collect themselves; pauses mid-sentence and clears throat; Pauses excessively; Stops mid-sentence; the way she pauses mid-sentence to collect her thoughts; Pauses before answering questions, as if weighing safety; pausing before answering questions; Pausing for longer than usual; pauses mid-sentence to glance around the room \\
\addlinespace[2pt]
Types of smiles & 222  & smiling too broadly, almost; smiles too brightly; Her smile feels like a thin, brittle thing, barely holding; Their gaze lingers on a photograph, a ghost of a smile touching their lips; Forced smiles that don't reach the eyes; smirks too brightly; A small, satisfied smile creeps onto their face; smiling tightly to avoid direct eye contact; Their smile falters; forced smile \\
\addlinespace[2pt]
Manners of breathing & 181  & took deep breaths; Inhales through the nose; breath coming in short, shallow gasps; makes a sharp breathing sound; Their breath quickens, taking short, shallow sips of air as excitement mounts; Takes deep, controlled breaths; their chewing slows or their hand stills mid-task for exactly one breath; takes a deep breath and releases it sharply; pauses before speaking, taking a deep breath; taking a long, slow breath \\
\addlinespace[2pt]
Fidgeting with small objects  & 165  & begin fidgeting with a ring, watch strap, or sleeve cuff in a repetitive, self-soothing rhythm; Fidgeting with a keychain or other small object in their pocket; Constant fidgeting; Fidgeting with a small object, like a pen cap or a ring, twisting it repeatedly; fidgeting with hands in pockets; A constant, restless fidgeting with a necklace or ring; fidgets nervously, adjusting position frequently; fidgeting with a pen or keychain; Sudden, unexplained fidgeting, particularly with hands or clothing; fidgeting with a pen on a cluttered desk \\
\addlinespace[2pt]
Fidgets with small objects  & 160  & fidgets with a small, portable device (e.g. phone, compass); fidgets with pockets; Fidgets with a necklace or ring; Fidgets with a ring or piece of jewelry associated with past failure; fidgets with wallet or purse; fidgets with a button or a small object in their pocket; Fidgets with a small object, like a pen cap or ring; Fidgets with a small object, like a pen or a keychain; fidgets with objects, rearranging them in a precise, calculated pattern; Fidgets with a piece of jewelry \\
\addlinespace[2pt]
Manners of leaning & 142  & leans away subtly; leaning in; Leaning forward aggressively; Leaning back, showing satisfaction or tiredness; Leans forward or forward-leans over objects to get closer to a goal; leans away from group conversations; leaning on a table or chair; Leaning forward in their chair, elbows on knees; Leans forward; leans back in a relaxed, trusting posture, indicating comfort and ease \\
\addlinespace[2pt]
Taps foot & 142  & Taps foot rhythmically, a sign of inner agitation; taps her foot rapidly, a hidden agitation; Tapping her left wrist absent-mindedly; tapping foot on the floor; Taps their foot impatiently; taps their foot rhythmically in moments of silence; tapping foot on floor; Taps fingers or foot nervously; taps their foot rapidly under the table; taps feet impatiently \\
\addlinespace[2pt]
Checks phone or watch & 135  & Regularly checks their phone for messages or alerts; a recurring habit of checking their watch or phone; Constantly checks their phone for news updates, even in the middle of conversations; Frequently checks their phone for messages from friends or family; Checks phone or looks at watch often; always checking the other person's phone for messages; frequently checks their phone for messages from a trusted friend; Frequently checks her watch or phone; Distractedly checks their phone; Subconsciously checks their watch or phone frequently \\
\addlinespace[2pt]
Crosses arms & 131  & Crossing her arms over her chest; Holding arms up slightly; Crossing arms or fingers in front; Crossing arms over chest; Crossed arms indicating tension or unresolved conflict; folds arms across chest; wraps arms tightly around themselves; Crossing their arms to block the view; crosses arms tightly; Crosses arms and legs tightly \\
\bottomrule
\end{tabular}
\caption{The ten largest indicator phrase clusters, showing the total number of phrases in the cluster and ten randomly sampled distinct phrases from the cluster (after cleaning and parsing)}
\label{tab:cluster-examples}
\end{table*}

\begin{table*}[t]
\centering\small
\setlength{\tabcolsep}{4pt}
\renewcommand{\arraystretch}{1.15}
\begin{tabular}{@{}>{\raggedright\arraybackslash}p{0.12\textwidth}>{\raggedright\arraybackslash}p{0.196\textwidth}>{\raggedright\arraybackslash}p{0.196\textwidth}>{\raggedright\arraybackslash}p{0.196\textwidth}>{\raggedright\arraybackslash}p{0.196\textwidth}@{}}
\toprule
\textbf{Model} & \cat{wound}{\textbf{Wound}:} \textit{being fired or laid off} & \cat{emotion}{\textbf{Emotion}:} \textit{guilt} & \cat{fear}{\textbf{Fear}:} \textit{crowds} & \cat{motivation}{\textbf{Motivation}:} \textit{being accepted by others} \\
\midrule
Claude Sonnet 4.6 & Keeps their work badge clipped to their jacket even on weekends, touching it reflexively before remembering & Volunteers for the most unpleasant tasks before anyone asks, almost punishingly so & They position themselves with their back against a wall or pillar before doing anything else in the space & mirrors the posture and hand gestures of whoever they're speaking with, shifting again each time they move to a new person in the group \\
\midrule
Llama-3.2-1B & shaking hands quickly when asked about the company & nervously fidgeting with hands & twitching nose & smiling wryly when someone compliments their work \\
Llama-3.2-3B & Fidgets with papers on desk & fidgeting with loose change & twitching nose in crowded spaces & fidgeting with a business card in their pocket \\
Llama-3.1-8B & Fidgets with the worn label on a old button-down shirt & fidgets with the hem of their shirt & Fidgeting with the hem of their shirt in public places & adjusting her outfit before meeting new people \\
Llama-3.1-70B & Avoids checking email or phone for messages from former employer & avoiding direct eye contact with a specific person or object & Avoids eye contact with strangers on public transportation & Mirrors the posture and gestures of those around them \\
Llama-3.3-70B & avoids eye contact during conversations about job security & avoids eye contact with specific individuals & avoids eye contact in public & mirrors others' mannerisms \\
Qwen-2.5-7B & Tightens collar at neck when sitting still for long periods & Fidgets with wedding ring on thumb while speaking & nervously adjusts glasses when near a large group & often mirrors the body language of those around them \\
Qwen-2.5-72B & flinches when a phone rings & Fidgets with a ring on their finger, avoiding eye contact. & Tenses shoulders and rolls them back when entering a busy room & leans in closer during group conversations \\
Gemma-2-9B & Avoids eye contact during conversations about work. & keeps tracing the rim of their coffee cup & constantly glances over shoulder & Mirroring the posture of the person they're speaking to. \\
Gemma-2-27B & Avoids eye contact during conversations about work & Avoids eye contact when the subject arises & Eyes constantly scanning the exits & Constantly seeks out opportunities to help others, even in small ways \\
Mistral-7B & Avoids eye contact during meetings, especially with their former supervisor & Frequent self-touching, particularly the collar or neck & Character maintains a personal space larger than usual when standing among people. & Excessive smiling, even in challenging situations \\
Phi-4 & meticulously reorganizes their desk and work area multiple times a day & Glances at the floor while speaking to others & glances repeatedly at exits and entry points, eyes darting & mirrors their group's posture and gestures subtly during conversations \\
R1-Distill-Qwen-7B & Avoiding eye contact & She fidgets with her hands and clutches at her skirt to avoid eye contact. & Slouching in a chair to avoid eye contact & slouching slightly when sitting \\
R1-Distill-Llama-8B & avoiding eye contact with coworkers, shifting gaze frequently & avoiding eye contact & Shifts weight uncomfortably, continuously adjusting position & adjusting their posture when someone nearby praises them \\
\bottomrule
\end{tabular}
\caption{The first indicator phrase each model generated for example states of each type (after cleaning and parsing).}
\label{tab:phrase-examples}
\end{table*}

\end{document}